\documentclass[11pt]{article}
\providecommand\JournalTitle[1]{#1}

\usepackage[colorlinks=true, allcolors=blue]{hyperref}

\usepackage{abstract, algorithm, algpseudocode, amsmath, amsthm, amssymb, appendix, bm, booktabs, caption, cite, etoolbox, lettrine, listofitems, listings, lmodern, fancyhdr, graphicx, hyperref, multirow, multicol, neuralnetwork, subfigure, tikz, titlesec, titling, wrapfig, xcolor}
\usepackage[subfigure]{tocloft}
\newcommand{\dropcap}[1]{\lettrine[lines=2,lraise=0.05,findent=0.1em, nindent=0em]{{\sffamily{#1}}}{}}

\usepackage[english]{babel}
\usetikzlibrary{shapes.geometric, arrows, bending}

\definecolor{gray}{cmyk}{0.05,0.1,0.1,0}

\usepackage[margin=0.5in, top=1.2in, bottom=1in]{geometry}
\pretitle{
    \begin{center}
    \vspace{-15mm} 
    \Huge
    \bfseries 
    \sffamily
} 
\posttitle{
    \end{center}
    \vspace{1mm}
} 

\renewcommand\thesection{\Roman{section}} 
\renewcommand\thesubsection{\roman{subsection}} 
\titleformat{\section}[block]{\Large\sffamily\bfseries}{\thesection.}{1em}{} 
\titleformat{\subsection}[block]{\large\bfseries\sffamily}{\thesubsection.}{1em}{} 

\title{Inferring networks from time series: \\ a neural approach} 

\date{}

\author{
	\bfseries{Thomas Gaskin$^{[1, 2, \star]}$}  
\and
	\bfseries{Grigorios A. Pavliotis$^{[2]}$} 
\and 
	\bfseries{Mark Girolami$^{[3, 4]}$}
}

\usepackage[scaled]{helvet}
\usepackage[T1]{fontenc}
\DeclareCaptionFormat{pnasformat}{\normalfont\sffamily\fontsize{8}{10}\selectfont#1#2#3}
\begin{document}
\newcommand{\cs}[1]{\sffamily\fontsize{8}{10}\selectfont{#1}}
\newcommand{\pred}[1]{\mathbf{\hat{#1}}}
\newcommand{\predm}[1]{\boldsymbol{\hat{#1}}}
\captionsetup{labelfont=bf}

\maketitle

\vspace{-10mm}
\begin{changemargin}{27pt}{27pt}
{\scriptsize\centering
$^\mathbf{1}$Department of Applied Mathematics and Theoretical Physics, University of Cambridge, Cambridge CB3 0WA, United Kingdom; $^\mathbf{2}$Department of Mathematics, Imperial College London, London SW7 2AZ, United Kingdom; $^\mathbf{3}$Department of Engineering, University of Cambridge, Cambridge CB2 1PZ, United Kingdom; $^\mathbf{4}$The Alan Turing Institute, London NW1 2DB, United Kingdom 

\medskip $^\star$To whom correspondence should be addressed: trg34@cam.ac.uk

}
\end{changemargin}

\vspace{6mm}
	
\begin{abstract} 

\noindent Network structures underlie the dynamics of many complex phenomena, from gene regulation and foodwebs to power grids and social media. Yet, as they often cannot be observed directly, their connectivities must be inferred from observations of the dynamics to which they give rise. In this work we present a powerful computational method to infer large network adjacency matrices from time series data using a neural network, in order to provide uncertainty quantification on the prediction in a manner that reflects both the degree to which the inference problem is underdetermined as well as the noise on the data. This is a feature that other approaches have hitherto been lacking. We demonstrate our method's capabilities by inferring line failure locations in the British power grid from its response to a power cut, providing probability densities on each edge and allowing the use of hypothesis testing to make meaningful probabilistic statements about the location of the cut. Our method is significantly more accurate than both Markov-chain Monte Carlo sampling and least squares regression on noisy data and when the problem is underdetermined, while naturally extending to the case of non-linear dynamics, which we demonstrate by learning an entire cost matrix for a non-linear model of economic activity in Greater London.  Not having been specifically engineered for network inference, this method in fact represents a general parameter estimation scheme that is applicable to any high-dimensional parameter space.
\bigskip

\noindent \textbf{Keywords}: Network inference, Neural differential equations, Model calibration, Power grids.
\end{abstract}

\vspace{6mm}
\hrule
\vspace{1mm}
\setcounter{tocdepth}{1}
\begin{multicols}{2}
  \tableofcontents
\end{multicols}\vspace{4mm}
\vspace{4mm}

\newpage

\twocolumn

\hrule height 0.1cm 
\section{Introduction}
\vspace{2cm}

\dropcap{N}etworks are important objects of study across the scientific disciplines. They materialise as physical connections in the natural world, for instance as the mycorrhizal connections between fungi and root networks that transport nutrients and warning signals between plants \cite{Simard_2012, Hettenhausen_2017}, human traffic networks \cite{Brockmann2013, Molkenthin2020}, or electricity grids \cite{Simonsen_2008, Shandilya_Timme_2011}. However, they also appear as abstract, non-physical entities, such as when describing biological interaction networks and food webs \cite{Stelzl2005, Proulx2005, Allesina2008}, gene or protein networks \cite{Tegner_2003, Palsson_2006, Sarmah_2022, Shen_2023}, economic cost relations \cite{Batty_2021, Ellam_2018}, or social links between people along which information (and misinformation) can flow \cite{DelVicario2016, Aral2013, Vosoughi2018}. In all examples, though the links constituting the network may not be tangible, the mathematical description is the same. In this work, we are concerned with inferring the structure of a static network from observations of dynamics on it. The problem is of great scientific bearing: for instance, one may wish to understand the topology of an online social network from observing how information is passed through it, and some work has been done on this question \cite{Myers_Leskovec_2010, Gomez_Rodriguez_2011, Gomez_Rodriguez_2012}. Another important application is inferring the connectivity of neurons in the brain by observing their responses to external stimuli \cite{Makarov_2005, Van_Bussel_2011}. In an entirely different setting, networks crop up in statistics in the form of conditional independence graphs, describing dependencies between different variables, which again are to be inferred from data \cite{Meinshausen_Buehlmann_2006, Yuan_Lin_2007}. 

Our approach allows inferring network connectivities from time series data with uncertainty quantification. Uncertainty quantification for network inference is important for two reasons: first, the observations will often be noisy, and one would like the uncertainty on the data to translate to the uncertainty on the predicted network. Secondly however, completely inferring large networks requires equally large amounts of data -- typically at least $N-1$ equations per node, $N$ being the number of nodes  -- and these observations must furthermore be linearly independent. Data of such quality and quantity will often not be available, leading to an underdetermined inference problem. The uncertainty on the predicted network should thus also reflect (at least to a certain degree) the `non-convexity' of the loss function, i.e. how many networks are compatible with the observed data. To the best of our knowledge, no current network inference method is able to provide this information.

Network inference can be performed using ordinary least squares (OLS) regression \cite{Shandilya_Timme_2011, Timme_2014}, but this is confined to the case where the dynamics are linear in the adjacency matrix. An alternative are sampling-based methods that generalise to the non-linear case \cite{Girolami-Calderhead-2011, Li_2016, Titsias_2023}, but these tend to struggle in very high-dimensional settings and can be computationally expensive. Efficient inference methods for large networks have been developed for cascading dynamics \cite{Myers_Leskovec_2010, Gomez_Rodriguez_2011, Gomez_Rodriguez_2012}, but these are highly specialised to a particular type of observation data and give no uncertainty quantification on the network prediction. Our method avoids these limitations. Its use of neural networks is motivated by their recent and successful application to low-dimensional parameter calibration problems \cite{Goettlich_2021, Gaskin_2023}, both on synthetic and real data, as well as by their conceptual proximity to Bayesian inference, e.g. through neural network Gaussian processes or Bayesian neural networks \cite{Lee_2018, Matthews_2018, Novak_2019, Kingma_Welling_2013, Blundell_2015, Gal_Ghahramani_2016}. Our method's underlying approach ties into this connection, and in fact, since it has not been specifically engineered to fit the network case, constitutes a general and versatile parameter estimation method. 

\paragraph{Method description} We apply the method proposed in \cite{Gaskin_2023} to the network case. The approach consists of training a neural network to find a graph adjacency matrix $\pred{A} \in \mathbb{R}^{N \times N}$ that, when inserted into the model equations, reproduces the observed time series $\mathbf{T} = (\mathbf{x}_1, ..., \mathbf{x}_L)$. A neural network is a function $u_\theta: \mathbb{R}^{N \times q} \to \mathbb{R}^p$, where $q \geq 1$ represents the number of time series steps that are passed as input. Its output is the (vectorised) estimated adjacency matrix $\pred{A}$, which is used to run a numerical solver for $B$ iterations ($B$ is the batch size) to produce an estimated time series $\mathbf{\hat{T}}(\pred{A}) = (\pred{x}_i, ..., \pred{x}_{i+B})$. This in turn is used to train the internal parameters $\boldsymbol{\theta}$ of the neural net (the weights and biases) via a loss function $J\left(\pred{A} \;\middle|\; \mathbf{T}\right)$.  The likelihood of any sampled estimate is simply proportional to
\begin{equation}
    p\left(\pred{A} \;\middle\vert\; \mathbf{T} \right) \propto e^{-J},
\end{equation}
and by Bayes' rule, the posterior density is then
\begin{equation}
	\pi\left(\pred{A} \;\middle\vert\; \mathbf{T}\right) = p\left(\pred{A} \;\middle\vert\; \mathbf{T}\right) \times \pi^0(\pred{A}) 
\end{equation}
with $\pi^0$ the prior density \cite{Stuart_2010}. As $\pred{A} = \pred{A}(\boldsymbol{\theta})$, we may calculate the gradient $\nabla_{\boldsymbol{\theta}}J$ and use it to optimise the internal parameters of the neural net using a backpropagation method of choice; popular choices include stochastic gradient descent, Nesterov schemes, or the Adam optimizer \cite{Kingma_2014}. Calculating $\nabla_{\boldsymbol{\theta}}J$ thus requires differentiating the predicted time series $\mathbf{\hat{T}}$, and thereby the system equations, with respect to $\pred{A}$. In other words: the loss function contains knowledge of the dynamics of the model. Finally, the true data is once again input to the neural net to produce a new parameter estimate $\pred{A}$, and the cycle starts afresh. A single pass over the entire dataset is called an epoch. 

Using a neural net allows us to exploit the fact that, as the net trains, it traverses the parameter space, calculating a loss at each point. Unlike Monte-Carlo sampling, the posterior density is not constructed from the frequency with which each point is sampled, but rather calculated directly from the loss value at each sample point. This entirely eliminates the need for rejection sampling or a burn-in: at each point, the true value of the likelihood is obtained, and sampling a single point multiple times provides no additional information, leading to a significant improvement in computational speed. Since the stochastic sampling process is entirely gradient-driven, the regions of high probability are typically found much more rapidly than with a random sampler, leading to a high sample density around the modes of the target distribution. We thus track the neural network's path through the parameter space and gather the loss values along the way. Multiple training runs can be performed in parallel, and each chain terminated once it reaches a stable minimum, increasing the sampling density on the domain, and ensuring convergence to the posterior distribution in the limit of infinitely many chains.

We begin this article with two application studies: first, we infer locations of a line failure in the British power grid from observations of the network response to the cut; and secondly, we infer economic cost relations between retail centres in Greater London. Thereafter we conduct a comparative analysis of our method's performance, before finally demonstrating the connection between the uncertainty on the neural net prediction and the uncertainty of the inference problem.

\vspace{0.5cm}
\hrule height 0.1cm

\section[Inferring line failures in the British power grid]{Inferring line failures in the \\ British power grid}

Power grids can be modelled as networks of coupled oscillators using the Kuramoto model of synchronised oscillation \cite{Kuramoto_1975, Filatrella_2008, Rohden_Timme_2012, Nishikawa_2015, Choi_2019}. Each node $i$ in the network either produces or consumes electrical power $P_i$ while oscillating at the grid reference frequency $\Omega$. The nodes are connected through a weighted undirected network $\mathbf{A} = (a_{ij})$, where the link weights $a_{ij} \sim Y_{ij} U_{ij}^2$ are obtained from the electrical admittances $Y_{ij}$ and the voltages $U_{ij}$ of the lines. The network coupling allows the phases $\varphi_i(t)$ of the nodes to synchronise according to the differential equation \cite{Nishikawa_2015}
\begin{equation}
	\alpha \dfrac{\mathrm{d}^2 \varphi_i}{\mathrm{d} t^2} + \beta \dfrac{\mathrm{d} \varphi_i}{\mathrm{d} t} = P_i + \kappa \sum_j a_{ij} \sin(\varphi_j - \varphi_i),
	\label{eq:second_order_Kuramoto}
\end{equation} 
where $\alpha$, $\beta$, and $\kappa$ are the inertia, friction, and coupling coefficients respectively. A requirement for dynamical stability of the grid is that $\sum_i P_i = 0$, i.e. that as much power is put into the grid as is taken out through consumption and energy dissipation \cite{Rohden_Timme_2012}. 

\begin{figure}[t!]
\begin{minipage}{0.5\textwidth}
	\cs{\textbf{(b)} Network response to line failure}
\end{minipage}	
\begin{minipage}{0.5\textwidth}
	\includegraphics[width=\textwidth]{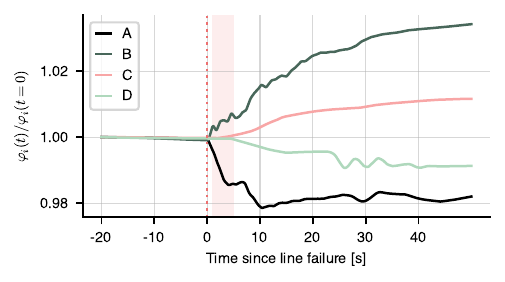}
\end{minipage}
	\caption{\textbf{(a)} Approximate high-voltage electricity transmission grid of Great Britain. Shown are 630 accurately placed nodes, representing power stations, substations, and transmission line intersections, and their connectivity as of January 2023 \cite{National_Grid_2023, SP_energy_networks_23, SSE_23}. Colours indicate the operating voltage of the lines. The size of the nodes indicate their power generation or consumption capacity (absolute values shown). White ringed nodes indicate the 38 nodes that are real power stations with capacities over 400 MW \cite{DUKES_2022}, with all other nodes assigned a random capacity in $[-200, +200]$. The two dotted edges in the northeast of England are the edges affected by a simulated power cut, labelled by the indices of their start and end vertices. \textbf{(b)} The network response to the simulated power line failure, measured at four different nodes in the network (marked A--D). The equation parameters were tuned to ensure phase-locking of the oscillators ($\alpha = 1$, $\beta=0.2$, $\kappa=30$). Nodes closer to the location of the line cut (A and B) show a stronger and more immediate response than nodes further away (C and D). The shaded area indicates the 4-second window we use to infer the line location.}
	\label{fig:power_grid}
\end{figure}

A power line failure causes the network to redistribute the power loads, causing an adjustment cascade to ripple through the network until equilibrium is restored \cite{Simonsen_2008}. In this work we recover the location of a line failure in the British power grid from observing these response dynamics. Figure \ref{fig:power_grid}a shows the high-voltage transmission grid of Great Britain as of January 2023, totalling 630 nodes (representing power stations, substations, and transformers) and 763 edges with their operating voltages. Of the roughly 1300 power stations dotted around the island, we include those 38 with installed capacities of at least 400 MW that are directly connected to the national grid \cite{DUKES_2022}; following \cite{Simonsen_2008, Rohden_Timme_2012} we give all other nodes a random value $P_i \sim \mathcal{U}[-200, +200]$ such that $\sum_i P_i =0$. 

We simulate a power cut in the northeast of England by iterating the Kuramoto dynamics until the system reaches a steady state of equilibrium (defined as $\vert \dot{\varphi}_i \vert/\varphi_i \leq 0.01 \ \forall i$) and then removing two links and recording the network response (fig. \ref{fig:power_grid}b). From the response we can infer the adjacency matrix of the perturbed network $\mathbf{\tilde{A}}$ (with missing links) and, by comparing with the unperturbed network $\mathbf{A}^0$ (without missing links), the line failure locations. 

\begin{figure}[t!]
\begin{minipage}{0.5\textwidth}
	\cs{\textbf{(a)} Power grid topology}
\end{minipage}	
\begin{minipage}{3.46in}
	\includegraphics[width=\textwidth]{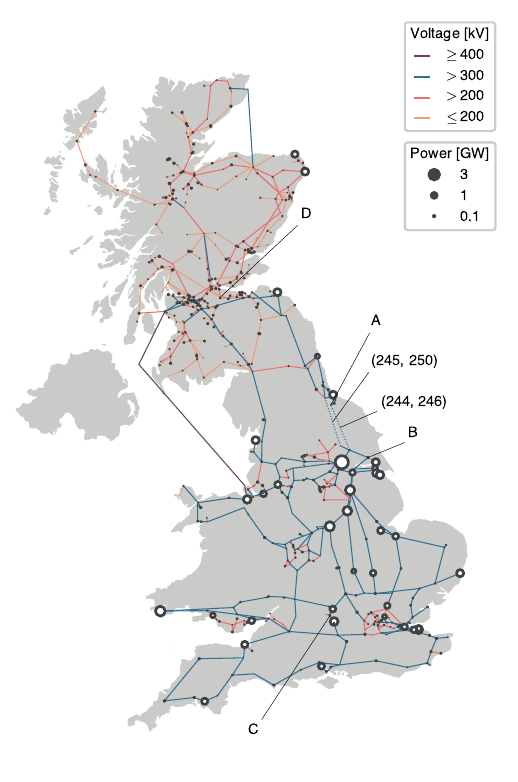}
\end{minipage}
\end{figure}

\begin{figure}[b!]
\includegraphics[width=0.5\textwidth]{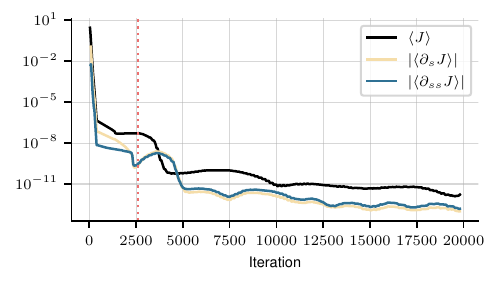}
\caption{The total loss $J$ and its derivatives with respect to the iteration count $\partial_s J$ and $\partial_{ss}J$, averaged over a window of 20 iterations (absolute values shown). The red dotted line indicates the value at which $\nu$ is set to 0.} 
\label{fig:cutoffs}
\end{figure}

\begin{figure*}[ht!]
\begin{minipage}{\textwidth}
	\cs{\textbf{(a)} Densities on edges with highest relative error}
\end{minipage}
 	\begin{minipage}{0.2457\textwidth}
 	 		\includegraphics[width=\textwidth]{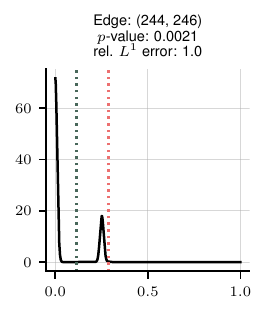}
 	\end{minipage}
 	 \begin{minipage}{0.2457\textwidth}
 	 		\includegraphics[width=\textwidth]{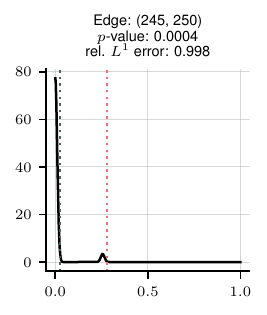}
 	\end{minipage}
 	 	\begin{minipage}{0.247\textwidth}
 	 		\includegraphics[width=\textwidth]{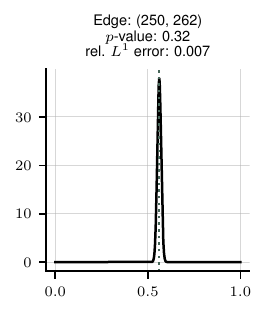}
 	\end{minipage}
 	 \begin{minipage}{0.247\textwidth}
 	 		\includegraphics[width=\textwidth]{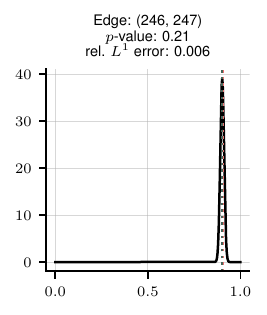}
 	\end{minipage}\hfill \vspace{4mm}
  \begin{minipage}{\textwidth}
	\cs{\textbf{(b)} True (black) and predicted network response}
\end{minipage}
    \begin{minipage}{\textwidth}
    		\includegraphics[width=\textwidth]{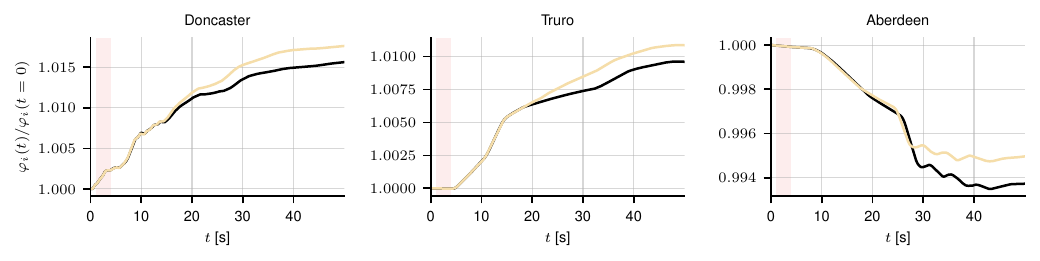}
    \end{minipage}
    \caption{Estimating the line failure location. \textbf{(a)} The densities on four edges with the highest relative prediction error $\vert \tilde{a}_{ij} - a^0_{ij} \vert/a^0_{ij}$ and their respective $p$-values for measuring the unperturbed value $a^0_{ij}$ ($\tilde{a}_{ij}$ is the prediction mode). Red dotted lines indicate the values of the unperturbed network, green lines the expectation values of the distributions. The marginals are smoothed using a Gaussian kernel.  We use a training set of length $L=400$ steps, and the batch size is $B=2$. CPU runtime: 24 minutes. \textbf{(b)} True (black) and predicted network responses at three different locations in the network. The responses are each normalised to the value at $t=0$. The shaded area represents the 400 time steps used to train the model. While the model is able to perfectly fit the response within the training range, it is not able to learn the full network from insufficient data, causing the time series to diverge for larger $t$.}
    	\label{fig:power_grid_results}
\end{figure*}

We let a neural network output a (vectorised) adjacency matrix $\pred{A}$ and use this estimated adjacency matrix to run the differential equation [\ref{eq:second_order_Kuramoto}], which will produce an estimate $\pred{T}$ of the observed time series of phases $\mathbf{T}$. A hyperparameter sweep on synthetic data showed that using a deep neural network with 5 layers, 20 nodes per layer, and no bias yields optimal results (see figs. \ref{fig:hyperparameters_1}—\ref{fig:hyperparameters_3} in the appendix). We use the hyperbolic tangent as an activation function on each layer except the last, where we use the `hard sigmoid' \cite{pytorch_hardsigmoid, tensorflow_hardsigmoid}
\begin{equation*}
	\sigma(x) = \begin{cases}
		0, \ x \leq -3, \\
		1, \ x \geq +3, \\
		x/6 + 1/2, \ \mathrm{else},
	\end{cases}\
\end{equation*}
which allows neural net output components to actually become zero, and not just asymptotically close, thereby ensuring sparsity of the adjacency matrix -- a reasonable assumption given that the power grid is far from fully connected. We use the Adam optimizer \cite{Kingma_2014} with a learning rate of $0.002$ for the gradient descent step. Since the neural network outputs are in $[0, 1]$, we scale the network weights $a_{ij} \to \lambda a_{ij}$ such that $a_{ij} \in [0, 1]$, and absorb the scaling constant $\lambda$ into the coupling constant $\kappa$; see the Supplementary Information for details on the calculations. 

We use the following loss function to train the internal weights $\boldsymbol{\theta}$ of the neural network such that it will output an adjacency matrix that reproduces the observed data:
\begin{align*}
    J\left(\pred{A} \;\middle|\; \mathbf{T} \right) = & \Vert \pred{T}(\pred{A}) - \mathbf{T} \Vert_2^2 + \Vert \pred{A} - \pred{A}^\top \Vert_2^2 + \mathrm{tr}(\pred{A}) \nonumber \\
	& + \nu \Vert \pred{A} - \mathbf{A}^0 \Vert_2^2.
\end{align*}
The first summand is the data-model mismatch, the second penalises asymmetry to enforce undirectedness of the network, and the third sets the diagonal to zero (which cannot be inferred from the data, since all terms $\sin(\theta_j - \theta_i) = 0$ for $i=j$). $\nu = \nu(s)$ is a function of the iteration count $s$ designed to let the neural network search for $\tilde{\mathbf{A}}$ in the vicinity of $\mathbf{A}^0$, since we can assume a priori that the two will be similar in most entries. To this end we set $\nu = 10$ while the loss function has not yet reached a stable minimum, quantified by $\vert\langle \partial_s  J \rangle \vert > 10^{-10}$ and $\vert \langle \partial_{ss} J \rangle \vert > 10^{-10}$, and $\nu=0$ thereafter. Here, $\langle \cdot \rangle$ is a rolling average over a window of $20$ iterations, see fig. \ref{fig:cutoffs}. In other words, we push the neural network towards a stable minimum in the neighbourhood of $\mathbf{A}^0$ and, once the loss stabilises, permanently set $\nu = 0$.

In theory $L = N-1$ observations are needed to completely infer the network, though symmetries in the data usually mean $L > N$ is required in practice \cite{Basiri_2018}. In this experiment we purposefully underdetermine the problem by only using $L < N-1$ steps; additionally, we train the network on data recorded $1$ simulated second after the power cut, where many nodes will still be close to equilibrium. Though the neural network may be unable to completely infer the network, it can nevertheless produce a joint distribution on the network edge weights $p\left(\pred{A} \;\middle|\; \mathbf{T}\right)$, recorded during the training, that allows us to perform hypothesis testing on the line failure location.
The marginal likelihood on each network edge $\hat{a}_{ij}$ is given by
\begin{equation}
	\rho(\hat{a}_{ij} \;\vert\; \mathbf{T}) = \int p\left(\pred{A} \;\middle\vert\; \mathbf{T} \right) \mathrm{d}\pred{A}_{-ij} \times \pi^0(\hat{a}_{ij}),
 \label{eq:edge_marginals}
\end{equation}
where the ${-ij}$ subscript indicates we are omitting the $ij$-th component of $\pred{A}$ in the integration. We assume uniform priors $\pi^0$ on each edge. In high dimensions, calculating the joint of all network edge weights can become computationally infeasible, but we can circumvent this by instead considering the two-dimensional joint density of the edge weight under consideration and the likelihood, $p(\hat{a}_{ij}, e^{-J})$ and then integrating over the likelihood,
\begin{equation}
    \rho(\hat{a}_{ij} \;\vert\; \mathbf{T}) = \int p\left(\hat{a}_{ij},   e^{-J}\right) \mathrm{d}(e^{-J}). 
\end{equation}
We show the results in fig. \ref{fig:power_grid_results}. Given the marginal distributions $\rho\left(\hat{a}_{ij} \;\middle|\; \mathbf{T}\right)$ with modes $\tilde{a}_{ij}$, we plot the densities on the four network edges with the highest relative prediction error $\vert \tilde{a}_{ij} -a^0_{ij}\vert / a^0_{ij}$. The advantage of obtaining uncertainty quantification on the network is now immediately clear: even in the underdetermined case we are able to make meaningful statistical statements about the line failure location. We see that the missing edges consistently have the highest relative prediction errors, and that the $p$-values for measuring the unperturbed value $a^0_{ij}$ under the null $\hat{a}_{ij}$ are 0.2\% and 0.04\% respectively, while being statistically insignificant for all other edges. It is interesting to note that the other candidate locations are also within the vicinity of the line failure, though their predicted values are much closer to the unperturbed value. In fig. \ref{fig:power_grid_results}b, we see that the predicted network reproduces the response dynamics for the range covered by the training data when inserted into eq. [\ref{eq:second_order_Kuramoto}], but, since the problem was purposefully underdetermined, the errors in the prediction $\pred{A}$ cause the predicted and true time series to diverge for larger $t$. Densities on all 200.000 potential edges were obtained in about twenty minutes on a regular laptop CPU. 

\vspace{0.5cm}
\hrule height 0.1cm

\section[Inferring economic cost networks from noisy data]{Inferring economic cost networks \\ from noisy data}

\begin{figure*}[t!]
	\begin{minipage}{\textwidth}
	\cs\flushleft{\textbf{(a)} Model dynamics}\vfill \vspace{2mm}
		\includegraphics[width=\textwidth]{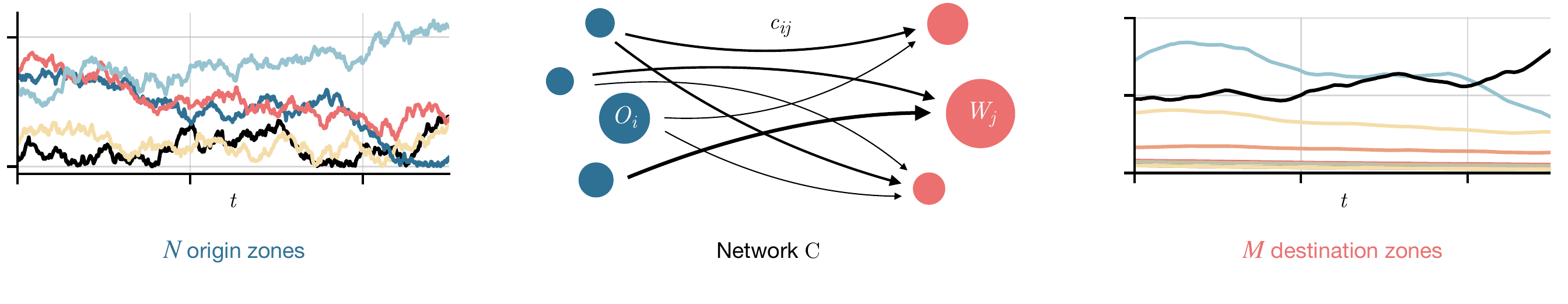}
	\end{minipage}\hfill\vspace{4mm}
	\begin{minipage}{\textwidth}
	\begin{minipage}[t]{0.5\textwidth} 
			\cs\flushleft{\textbf{(b)} Initial data and travel times network}	\vfill			\vspace{5mm}\centering	
			\includegraphics[width=0.9\textwidth]{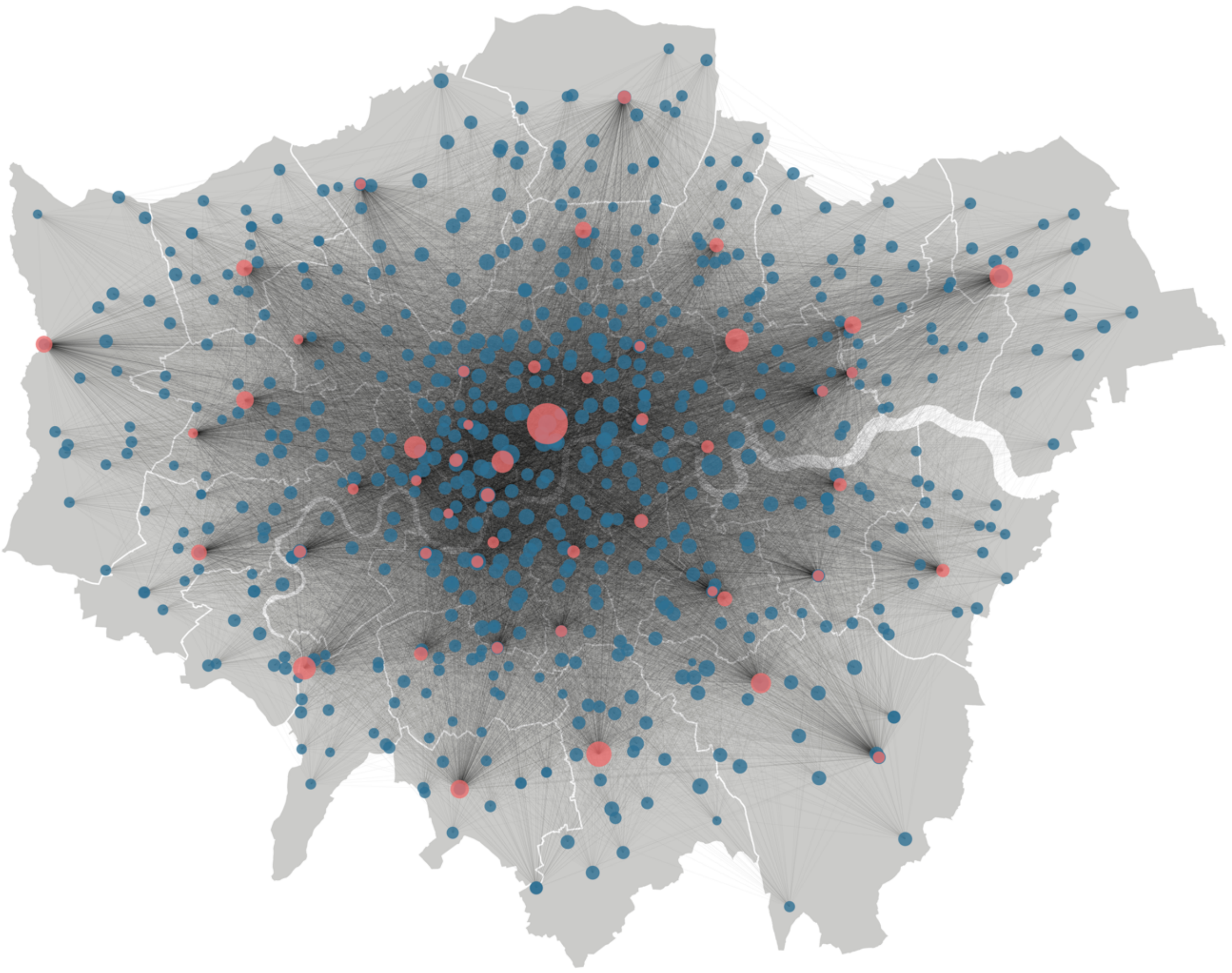}
		\end{minipage}
		\begin{minipage}[t]{0.5\textwidth} 
			\cs\flushleft{\textbf{(c)} Inferred weighted degree distribution}\vfill \vspace{5mm}
		\includegraphics[width=\textwidth]{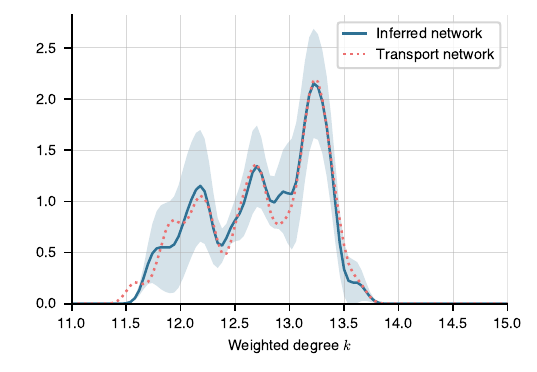}
		\end{minipage}
	\end{minipage}
	\caption{Inferring economic cost networks. \textbf{(a)} In the model, $N$ origin zones (red) are connected to $M$ destination zones (blue) through a weighted directed network. Economic demand flows from the origin zones to the destination zones, which supply the demand. We model the origin zones $O_i(t)$ as a Wiener process with diffusion coefficient $\sigma_O = 0.1$. The resulting cumulative demand at destination zone $j$ is given by $W_j$. Note that the origin zone sizes fluctuate more rapidly than the destination zones, since there is a delay in the destination zones' response to changing consumer patterns, controlled by the parameter $\epsilon$. We use the parameters as estimated in \cite{Gaskin_2023}, $\alpha=0.92$, $\beta=0.54$, $\kappa=8.3$, and set $\epsilon=2$. \textbf{(b)} The initial origin and destination zone sizes, given by the total household income of the $N=629$ wards in London (blue nodes) and the retail floor space of $M=49$ major centres (red nodes) \cite{GLA_health_check_report, GLA_ward_data}. The network is given by travel times as detailed in the text. Background map: \cite{GLA_GIS}. \textbf{(c)} Predicted degree distribution (sold line) of the inferred network, for a high noise level of $\sigma=0.14$, and one standard deviation (shaded area), and the true distribution (red dotted line). CPU runtime: 3 min 41 s.}
	\label{fig:Harris_Wilson_results}
	\end{figure*}
In the previous example the underlying network was a physical entity, but in many cases networks model abstract connections. We therefore now consider a commonly used economic model of the coupling of supply and demand \cite{HarrisWilson78, Ellam_2018, Batty_2021} and a dataset of economic activity across Greater London. The goal is to learn the entire coupling network, not just to infer the (non-)existence of individual edges. In the model, $N$ origin zones of sizes $O_i$, representing economic demand, are coupled to $M$ destination zones of sizes $W_j$, modelling the supply side, through a network whose weights quantify the convenience with which demand from zone $i$ can be supplied from zone $j$: the higher the weight, the more demand flows through that edge (see fig. \ref{fig:Harris_Wilson_results}a). Such a model is applicable e.g. to an urban setting \cite{Batty_2021}, the origin zones representing residential areas, the destination zones e.g. commercial centres, and the weights quantifying the connectivity between the two (transport times, distances, etc.). The resulting cumulative demand at destination zone $j$ depends both on the current size $W_j(t)$ of the destination zone and the network weights $c_{ij}$: 
\begin{equation}
	D_j = \sum_{i=1}^{N} \dfrac{W_j(t)^\alpha c_{ij}^\beta}{\sum_{k=1}^M W_k(t)^\alpha c_{ik}^\beta} O_i(t).
 \label{eq:HarrisWilson}
\end{equation}
The sizes $W_j$ are governed by a system of $M$ coupled logistic Stratonovich stochastic differential equations
\begin{equation}
	\mathrm{d}W_j = \epsilon W_j(D_j - \kappa W_j)\mathrm{d}t + \sigma W_j \circ \mathrm{d}\xi_j,
 \label{eq:HarrisWilsonSDE}
\end{equation}
with given initial conditions $W_j(0)$, see fig. \ref{fig:Harris_Wilson_results}a. $\alpha$, $\beta$, $\kappa$, and $\epsilon$ are scalar parameters. Our goal is to infer the cost matrix $\mathbf{C} = (c_{ij})$ from observations of the time series $\mathbf{O}(t)$ and $\mathbf{W}(t)$. The model includes multiplicative noise with strength $\sigma \geq 0$, where the $\xi_j$ are independent white noise processes and $\circ$ signifies Stratonovich integration \cite{Pavliotis_2014}. Crucially, the model depends non-linearly on $\mathbf{C}$.

We apply this model to a previously studied dataset of economic activity in Greater London \cite{Ellam_2018, Gaskin_2023}. We use the ward-level household income from $N=625$ wards for 2015 \cite{GLA_ward_data} and the retail floor space of the $M=49$ largest commercial centres in London \cite{GLA_health_check_report} as the initial origin zone and destination zone sizes respectively, i.e. $\mathbf{O}(0)$ and $\mathbf{W}(0)$, and from this generate a synthetic time series using the parameters estimated in \cite{Gaskin_2023} for a high noise level of $\sigma=0.14$. For the network $\mathbf{C}$ we use the Google Distance Matrix API\footnote{\href{https://developers.google.com/maps/documentation/distance-matrix}{developers.google.com/maps/documentation/distance-matrix}} to extract the shortest travel time $d_{ij}$ between nodes, using either public transport or driving. The network weights are derived in \cite{Wilson_1967} as 
\begin{equation*}
	c_{ij} = e^{- d_{ij}/\tau}
\end{equation*}
where the scale factor $\tau = \max_{i,j} d_{ij}$ ensures a unitless exponent.

 We generate a synthetic time series of $10000$ time steps, from which we subsample $2500$ 2-step windows, giving a total training set size of $L=5000$ time steps. This is to ensure we sample a sufficiently broad spectrum of the system's dynamics, thereby fully determining the inference problem and isolating the effect of the training noise. A hyperparameter sweep on synthetic data showed that using a neural network with $2$ layers, 20 nodes per layer, and no bias yields optimal results. We use the hyperbolic tangent as the activation function on all layers except the last, where we use the standard sigmoid function (since the network is complete, there is no need to use the hard sigmoid as all edge weights are nonzero). To train the neural network we use the simple loss function
\begin{equation*}
	J = \Vert \pred{T}(\pred{A}) - \mathbf{T} \Vert_2^2, 
 \end{equation*}
where $\pred{T}$ and $\mathbf{T}$ are the predicted and true time series of destination zone sizes. Since the dynamics are invariant under scaling of the cost matrix $\mathbf{C} \to \lambda \mathbf{C}$, we normalise the row sums of the predicted and true networks, $\sum_j c_{ij}=1$. 

Figure \ref{fig:Harris_Wilson_results}c shows the inferred distribution $\rho(k)$ of the (weighted) origin zone node degrees $k_j = \sum_i c_{ij}$. The solid line is the maximum likelihood prediction, and the dotted red line the true distribution . Even with a high level of noise, the model manages to accurately predict the underlying connectivity matrix, comprising over 30.000 weights, in under 5 minutes on a regular laptop CPU. Uncertainty on $P(k)$ is given by the standard deviation,
\begin{equation}
    \mathbb{E}_{\pred{T}} \left[P\left(k \;\middle|\; \pred{T}\right) - \hat{P}(k)\right]^2,
    \label{eq:std}
\end{equation}
where $\hat{P}$ is the maximum likelihood estimator. As we will discuss in the last section, this method meaningfully captures the uncertainty due to the noise in the data and the degree to which the problem is underdetermined.

\vspace{0.5cm}
\hrule height 0.1cm

\section{Comparative performance analysis}
We now analyse our method's performance, both in terms of prediction quality and computational speed, by comparing it to a Markov-Chain Monte Carlo approach (MCMC) as well as a classical regression method, presented e.g. in \cite{Timme_2007, Shandilya_Timme_2011}. As mentioned in the introduction, computationally efficient network learning methods have been developed for specific data structures; however, we compare our approach with MCMC and OLS since both are general in the types of data to which they are applicable.

Consider noisy Kuramoto dynamics,
\begin{equation}
	\alpha\dfrac{\mathrm{d}^2 \varphi_i}{\mathrm{d} t^2} + \dfrac{\mathrm{d} \varphi_i}{\mathrm{d} t} - \omega_i =  \sum_j a_{ij} \sin(\varphi_j - \varphi_i) + \xi_i,
	\label{eq:Kuramoto_model}
\end{equation}
with $\xi_i$ independent white noise processes with strength $\sigma$, and $\omega_i$ the eigenfrequencies of the nodes. Given $L$ observations of each node's dynamics, we can gather the left side into a single vector $\mathbf{X}_i \in \mathbb{R}^{1 \times L}$ for each node, and obtain $N$ equations
\begin{equation}
	\mathbf{X}_i = \mathbf{A}_i \cdot \mathbf{G}_i +\boldsymbol{\xi}_i, \ i = 1, ..., N,
\end{equation}
with $\mathbf{A}_i \in \mathbb{R}^{1 \times N}$ the $i$-th row of the adjacency matrix $\mathbf{A}$, and $\mathbf{G}_i \in \mathbb{R}^{N \times L}$ the $L$ observations of the the interaction terms $\sin(\varphi_j-\varphi_i), \ j=1, ..., N$. From this we can then naturally estimate the $i$-th row of $\mathbf{A}$ using ordinary least squares:
\begin{equation}
	\pred{A}_i = \underset{\boldsymbol{\gamma} \in \mathbb{R}^{1 \times N}}{\mathrm{argmin}} \ \Vert \mathbf{X}_i - \boldsymbol{\gamma} \cdot \mathbf{G}_i \Vert_2^2 = \mathbf{X}_i \mathbf{G}_i^\top \left( \mathbf{G}_i \mathbf{G}_i^\top \right)^{-1}.
	\label{eq:OLS_2}
\end{equation}
Given sufficiently many linearly independent observations, the Gram matrices $\mathbf{G}_i \mathbf{G}_i^\top$ will all be invertible; in the underdetermined case, a pseudoinverse can be used to approximate their inverses. As before, the diagonal of $\pred{A}$ is manually set to 0. 

In addition to regression, we also compare our method to a preconditioned Metropolis-adjusted Langevin sampling scheme (MALA) \cite{Girolami-Calderhead-2011, Li_2016, Chewi-2021, Titsias_2023}, which constructs a Markov chain of sampled adjacency matrices $\mathbf{\hat{A}}$ by drawing proposals from the normal distribution
\begin{equation}
    \mathbf{\hat{A}}^{i+1} \sim \mathcal{N} \left(\pred{A}^{i} + \dfrac{\tau}{2} \lambda^{-1}\mathbf{P} \nabla J\left(\pred{A}^i \;\vert\; \mathbf{T} \right), \tau \lambda^{-1} \mathbf{P} \right).
\end{equation}
Here, $\tau > 0$ is the integration step size, $\mathbf{P} \in \mathbb{R}^{N^2 \times N^2}$ a preconditioner (note that we are reshaping $\pred{A}$ into an $N^2$-dimensional vector), and $\lambda = \mathrm{tr}(\mathbf{P})/N^2$ its average eigenvalue. Each proposal is accepted with probability
\begin{equation}
    \eta = \min \left[1,  \dfrac{\exp(-J(\pred{A}^{i+1}))q\left(\pred{A}^{i+1} \;\middle|\; \pred{A}^{i}\right)}{
    \exp(-J(\pred{A}^{i})) q\left(\pred{A}^{i} \;\middle|\;\pred{A}^{i+1}\right)
    } \right],
\end{equation}
with the transition probability
\begin{equation}
    q\left(\pred{A}^{i+1} \;\middle|\;\pred{A}^i\right) \propto \exp\left({-\dfrac{1}{4\tau} \Vert \pred{A}^{i+1}} - \pred{A}^{i} - \tau \nabla \log \pi(\pred{A}^{i}) \Vert_2^2 \right).
\end{equation}
We tune $\tau$ so that the acceptance ratio $\eta$ converges to the optimum value of $0.57$ \cite{Roberts_Rosenthal_2002}.

We set the preconditioner $\mathbf{P}$ to be the inverse Fisher information covariance matrix
\begin{equation}
    \mathbf{P}^{-1} = \mathbb{E}_{\mathbf{\hat{A}}} \left[ \nabla J (\pred{A}^i) \nabla J(\pred{A}^i)^\top \right],
\end{equation}
which has been shown to optimise the expected squared jump distance \cite{Titsias_2023}. The expectation value is calculated empirically over all samples drawn using the efficient algorithm given in \cite{Titsias_2023}. In all experiments, we employ a `warm start' by initialising the sampler close to the minimum of the problem. We found this to be necessary in such high dimensions (between 256 and 490.000) to produce decent results. Unlike the MCMC sampler, the neural network is initialised randomly.

\begin{figure*}[ht!]
\begin{minipage}{0.33\textwidth}
	{\cs{\textbf{(a)} Accuracy as a function of noise}}
\end{minipage}
\begin{minipage}{0.33\textwidth}
	{\cs{\textbf{(b)} Accuracy as a function of convexity}}
\end{minipage}
\begin{minipage}{0.33\textwidth}
	{\cs{\textbf{(c)} Compute times}}
\end{minipage}
\includegraphics[width=0.33\textwidth]{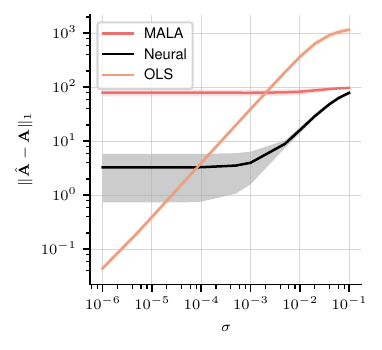}
\includegraphics[width=0.33\textwidth]{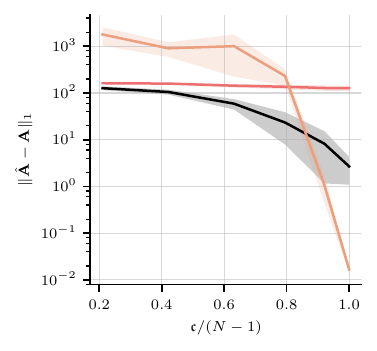}
\includegraphics[width=0.33\textwidth]{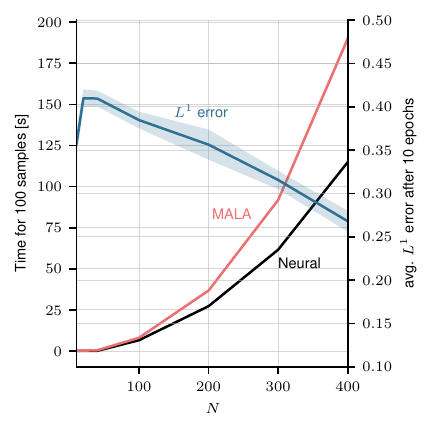}
	\begin{minipage}{0.5\textwidth}
	{\cs{ \textbf{(d)} Degree distribution, $N=1000$}}
	\end{minipage}
	\begin{minipage}{0.5\textwidth}
	{\cs{\textbf{(e)} Triangle distribution, $N=1000$}}
	\end{minipage}
	\begin{minipage}{0.5\textwidth}
		\includegraphics[width=\textwidth]{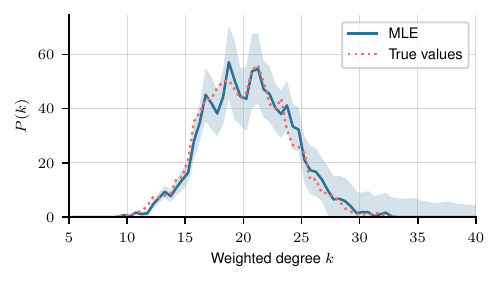}
	\end{minipage}
	\begin{minipage}{0.5\textwidth}
		\includegraphics[width=\textwidth]{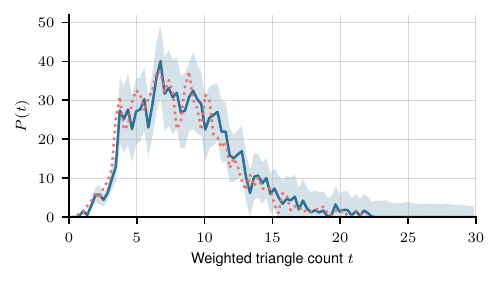}
	\end{minipage}
	\caption{Computational performance analysis. \textbf{(a)} $L^1$ prediction error eq. [\ref{eq:L1_error}] of the neural scheme, the preconditioned Metropolis-adjusted Langevin sampler, and OLS regression as a function of the noise variance $\sigma$ on the training data. For very high noise levels, the training data is essentially pure noise, and the prediction errors begin to plateau. First-order Kuramoto dynamics are used ($\alpha =0$), though these results also hold for second-order dynamics (cf. fig \ref{fig:second_order_accuracy} in the appendix). Enough data is used to ensure full invertibility of the Gram matrix ($\mathfrak{c}=1$). \textbf{(b)} The $L^1$ accuracy as a function of the convexity $\mathfrak{c}$ of the loss function (eq. [\ref{eq:convexity}]). \textbf{(c)} Compute times for ten epochs, or 100 samples, of the neural scheme and the preconditioned Metropolis-adjusted Langevin sampler (MALA), averaged over 10 runs. The shaded areas showing one standard deviation. On the right axis, the average $L^1$ prediction error of the neural scheme $\frac{1}{N} \Vert \pred{A} - \mathbf{A} \Vert_1$ after 10 epochs is shown, which remains fairly constant as a function of $N$, showing that the number of gradient descent steps required to achieve a given average prediction error does not depend on $N$. \textbf{(d)} Predicted degree distribution and \textbf{(e)} triangle distribution of an inferred network with $N=1000$ nodes, trained on first-order noisy Kuramoto data ($\sigma = 0.001$). The blue shaded areas indicate one standard deviation, and the red dotted lines are the true distributions. CPU runtime: 1 hour 3 minutes.}
	\label{fig:computational_performance}
\end{figure*}

Figures \ref{fig:computational_performance}a--b show our method's prediction accuracy alongside that of OLS regression and preconditioned MALA on synthetic Kuramoto data; the accuracy here is defined as the $L^1$ error
\begin{equation}
	\Vert \pred{A} - \mathbf{A} \Vert_1 = \sum_{i, j} \vert \hat{a}_{ij} - a_{ij} \vert,
 \label{eq:L1_error}
\end{equation}
where $\pred{A}$ is the mode of the posterior. In fig. \ref{fig:computational_performance}a, the accuracy is shown as a function of the noise $\sigma$ on the training data. We generate enough data to ensure the likelihood function is unimodal. For the practically noiseless case of $\sigma < 10^{-5}$, the regression scheme on average outperforms the neural approach; however, even for very low noise levels $\sigma \geq 10^{-5}$ and above, the neural approach proves far more robust, outperforming OLS by up to one order of magnitude and maintaining its prediction performance up to low noise levels of $\sigma \leq 10^{-3}$. Meanwhile, we find that in the low- to mid-level noise regime, the neural scheme approximates the mode of the distribution by between one to two orders of magnitude more accurately than the Langevin sampler. For high levels of noise ($\sigma > 10^{-2}$), the performances of the neural and MALA schemes converge. These results hold both for first-order ($\alpha=0$) and second-order Kuramoto dynamics [\ref{eq:second_order_Kuramoto}]; in the second-order case, the neural method begins outperforming OLS at even lower levels of $\sigma$ than in the first-order case, though the improvement is not as significant (cf fig. \ref{fig:second_order_accuracy} in the appendix).

In figure \ref{fig:computational_performance}b we show the accuracy as a function of the convexity of the loss function. In general, it is hard to quantify the convexity of $J$, since we do not know how many networks fit the equation at hand. However, when the dynamics are linear in the adjacency matrix $\mathbf{A}$, we can do so using the Gram matrices of the observations of each node $i$, $\mathbf{G}_i\mathbf{G}_i^\top$, where we quantify the (non-)convexity of the problem by the minimum rank of all the Gram matrices,
\begin{equation}
	\mathfrak{c} := \min_i \mathrm{rk}\left( \mathbf{G}_i\mathbf{G}_i^\top \right).
	\label{eq:convexity}
\end{equation}
The problem is fully determined if $\mathfrak{c} = N-1$ and all Gram matrices are invertible. As shown, regression is again more accurate when the problem is close to fully determined; however, as $\mathfrak{c}$ decreases, the accuracy quickly drops, with the neural scheme proving up to an order of magnitude more accurate. Meanwhile, the MCMC scheme is consistently outperformed by the neural scheme, though it too eclipses regression for $\mathfrak{c}<0.75$. In summary, regression is only viable for the virtually noiseless and fully determined case, while the neural scheme maintains good prediction performance even in the noisy and highly underdetermined case (see also fig. \ref{fig:computational_performance}d--e).
\begin{figure*}[t]
	\begin{minipage}{0.333\textwidth}
	\cs{\textbf{(a)} Convexity uncertainty on $P(k)$}
	\end{minipage}	
	\begin{minipage}{0.333\textwidth}
	\cs{\textbf{(b)} Noise uncertainty on $P(k)$}
	\end{minipage}	
	\begin{minipage}{0.333\textwidth}
	\cs{\textbf{(c)} Noise uncertainty on $P(t)$}
	\end{minipage}
    \begin{minipage}{\textwidth}
    \includegraphics[height=0.333\textwidth]{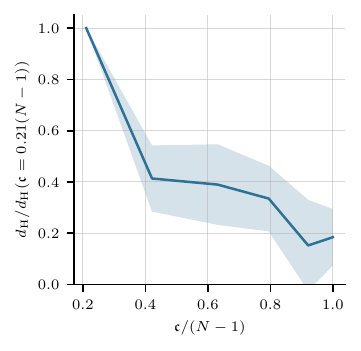}
    \includegraphics[height=0.333\textwidth]{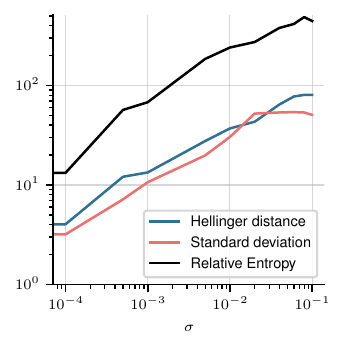}
    \includegraphics[height=0.333\textwidth]{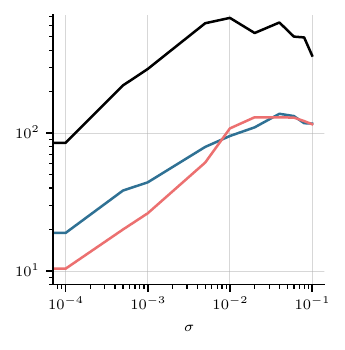}
    \end{minipage}
	\caption{Quantifying the two types of uncertainty:  \textbf{(a)} Hellinger error (eq. [\ref{eq:Hellinger_distance}]) on the degree distribution $P(k)$ as a function of $\mathfrak{c}$ (eq. [\ref{eq:convexity}]) in the noiseless case. The error is normalised to the value at $\mathfrak{c}=0.21(N-1)$. As $\mathfrak{c}$ increases, the error on the prediction decreases almost linearly. We run the model from 10 different initialisations and average over each (shaded area: standard deviation). \textbf{(b)} and \textbf{(c)}: Prediction uncertainty due to noise in the data. Shown are the expected Hellinger error (eq. [\ref{eq:Hellinger_distance}]) and expected relative entropy (eq. [\ref{eq:relative_entropy}]) to the maximum likelihood estimate, as well as the total standard deviation $s$, eq. [\ref{eq:std}], for the degree distribution $P(k)$ and triangle distribution $P(t)$ as a function of the noise $\sigma$ on the data. Each line is an average over 10 different initialisations. In all cases, training was conducted on synthetic, first-order Kuramoto data (eq. [\ref{eq:Kuramoto_model}] with $\alpha=0$).}
	\label{fig:Uncertainty}
\end{figure*}

In figure \ref{fig:computational_performance}c we show compute times to obtain 100 samples for both the neural and MALA schemes. The complexity of the neural scheme is 
$\mathcal{O}(n_\mathrm{E} \times L N^2)$, with $n_\mathrm{E}$ the number of training epochs. This is because each epoch of the model equation requires $\mathcal{O}(L N^2)$ operations for the vector-matrix multiplication in eq. [\ref{eq:OLS_2}], and $\mathcal{O}(LN^2/B)$ for the stochastic gradient descent update, where we are holding $L/B$ constant to ensure comparability. As is visible, the average $L^1$ error per edge weight remains constant over $N$, showing that the number of epochs required to achieve a given node-averaged prediction accuracy is independent of $N$. The preconditioned MALA scheme is considerably slower, due to the computational cost of calculating the preconditioner and the Metropolis-Hastings rejection step.

Lastly, figures \ref{fig:computational_performance}d--e show the estimated weighted degree and triangle distributions of a large graph with 1000 nodes, or 1 million edge weights to be estimated, for noisy training data. The number of weighted, undirected triangles on each node $i$ is given by $\frac{1}{2}\sum_{jk} a_{ij}a_{jk}a_{ki}$. The model robustly finds the true adjacency matrix, and we again quantify uncertainty on the prediction using the standard deviation eq. [\ref{eq:std}]. Estimating a network with 1000 nodes on a standard laptop CPU took about 1 hour, which reduces to 6 minutes when using a GPU. Most high-performance network inference techniques demonstrate their viability on graphs with at most this number of nodes, e.g. ConNIe \cite{Myers_Leskovec_2010} and NetINF \cite{Gomez_Rodriguez_2012}. In \cite{Myers_Leskovec_2010}, the authors state that graphs with 1000 nodes can typically be inferred from cascade data in under 10 minutes on a standard laptop. Similarly, the authors of NetINF \cite{Gomez_Rodriguez_2012} state that it can infer a network with 1000 nodes in a matter of minutes, though this algorithm does not infer edge weights, only the existence of edges, and neither technique provides uncertainty quantification.

\pagebreak
\hrule height 0.1cm
\section{Quantifying uncertainty}
There are two sources of uncertainty when inferring adjacency matrices: the non-convexity of the loss function $J$, and the noise $\sigma$ on the data.  In figure \ref{fig:Uncertainty}a we show the expected Hellinger error 
\begin{equation}
\dfrac{1}{2}\mathbb{E}_{\pred{T}} \int  \left[\sqrt{P\left(x \;\middle|\; \pred{T} \right)} - \sqrt{\hat{P}(x)}  \right]^2 \mathrm{d}x
	\label{eq:Hellinger_distance}
\end{equation}
on the predicted degree distribution as a function of $\mathfrak{c}$. As is visible, the error on the distribution decreases as $\mathfrak{c}$ tends to its maximum value of $N-1$. For $\mathfrak{c}=N-1$, some residual uncertainty remains due to the uncertainty on the neural network parameters $\boldsymbol{\theta}$.

In figures \ref{fig:Uncertainty}b--c we show the expected Hellinger error (eq. [\ref{eq:Hellinger_distance}]) on the maximum likelihood estimator $\hat{P}$ as a function of $\sigma$, for both the degree and triangle distributions, i.e. $x \in \{k, t\}$. In addition, we also show the behaviour of the expected relative entropy
\begin{equation}
   \mathbb{E}_{\pred{T}} \int  P\left(x \;\middle|\; \pred{T}\right) \log \left(\dfrac{P\left(x \;\middle|\; \pred{T}\right)}{\hat{P}(x)} \right) \mathrm{d}x
   \label{eq:relative_entropy}
\end{equation}
and the total standard deviation
\begin{equation}
    s^2 = \int \mathbb{E}_{\pred{T}} \left[P\left(x \;\middle|\; \pred{T}\right) - \hat{P}(x)\right]^2 \mathrm{d}x.
\end{equation}
All three metrics reflect the noise on the training data, providing similarly behaved, meaningful uncertainty quantification. As the noise tends to 0, some residual uncertainty again remains, while for very high noise levels the uncertainty begins to plateau. Our method thus manages to capture the uncertainty arising from both sources: the non-convexity of $J$ and the noise $\sigma$ on the data.

\pagebreak
\hrule height 0.1cm

\section{Discussion}In this work we have demonstrated a performative method to estimate network adjacency matrices from time series data. We showed its effectiveness at correctly and reliably inferring networks in a variety of scenarios: convex and non-convex cases, low to high noise regimes, and equations that are both linear and non-linear in $\mathbf{A}$. We were able to reliably infer power line failures in the national power grid of Great Britain, and the connectivity matrix of an economic system covering all of Greater London. We showed that our method is well able to handle inference of hundreds of thousands to a million edge weights, while simultaneously giving uncertainty quantification that meaningfully reflects both the non-convexity of the loss function as well as the noise on the training data. Our method is significantly more accurate than MCMC sampling, and outperforms OLS regression in all except the virtually noiseless and fully determined cases. This is an important improvement since large amounts of data are typically required to ensure the network inference problem is fully determined, which may often not be available, as suggested in the power grid study. Unlike regression, our method also naturally extends to the case of non-linear dynamics. In conjunction with our previous work \cite{Gaskin_2023}, we have now also demonstrated the viability of using neural networks for parameter calibration in both the low- and high-dimensional case. Our method is simple to implement as well as highly versatile, giving excellent results across a variety of problems. All experiments in this work were purposefully conducted on a standard laptop CPU, typically taking on the order of minutes to run. 

Many lines for future research open up from this work. Firstly, a thorough theoretical investigation of the method is warranted, establishing rigorous convergence guarantees and bounds on the error of the posterior estimate. Another direction is further reducing the amount of data required to accurately learn parameters, and in future research the authors aim to address the question of learning system properties from observations of a single particle trajectory at the mean-field limit \cite{Pavliotis_Zanoni_2022, Zagli_Pavliotis_2023}. In this work we have also not considered the impact of the network topology on the prediction performance, rather focusing on the physical dynamics of the problem. An interesting question is to what degree different network structures themselves are amenable to or hinder the learning process.

Over the past decade much work has been conducted into graph neural architectures \cite{Bronstein_2017, GNNBook2022}, the use of which may further expand the capabilities of our method. More specialised architectures may prove advantageous for different (and possibly more difficult) inference tasks, though we conducted a limited number of experiments with alternatives (e.g. autoencoders, cf. fig. \ref{fig:hyperparameters_3}) and were unable to find great performance improvements. Finally, one drawback of our proposed method in its current form is it that it requires differentiability of the model equations in the parameters to be learned; future research might aim to develop a variational approach to expand our method to weakly differentiable settings. 

\vspace{0.5cm}
\hrule 

\paragraph*{Data, materials, and Software Availability}
Code and synthetic data can be found under \url{https://github.com/ThGaskin/NeuralABM}. It is easily adaptable to new models and ideas. The code uses the \texttt{utopya} package\footnote{\href{https://utopia-project.org}{utopia-project.org}, \href{https://utopya.readthedocs.io/en/latest/}{utopya.readthedocs.io/en/latest}} \cite{Riedel2020, Sevinchan_2020} to handle simulation configuration and efficiently read, write, analyse, and evaluate data. This means that the model can be run by modifying simple and intuitive configuration files, without touching code. Multiple training runs and parameter sweeps are automatically parallelised. The neural core is implemented using \texttt{pytorch}\footnote{\href{https://pytorch.org}{pytorch.org}}. All synthetic datasets as well as the London dataset have been made available, together with the configuration files needed to reproduce the plots. Detailed instructions are provided in the supplementary material and the repository. The British power grid data \cite{National_Grid_2023, SP_energy_networks_23, SSE_23} is property of the respective organisations and cannot be made available without permission; however, as of early 2023 it is freely available from those organisations upon request. The code used to run the experiments is available in the repository.

\vspace{0.5cm}
\hrule 

\paragraph*{Author contributions} TG, GP, MG designed and performed the research and wrote the paper; TG wrote the code and performed the numerical experiments.

\vspace{0.5cm}
\hrule 

\paragraph*{Acknowledgements}{The authors are grateful to Dr Andrew Duncan (Imperial College London) for fruitful discussions on power grid dynamics, and to the anonymous reviewers for their helpful comments during the peer review process. TG was funded by the University of Cambridge School of Physical Sciences VC Award via DAMTP and the Department of Engineering, and supported by EPSRC grants EP/P020720/2 and \\ EP/R018413/2. The work of GP was partially funded by EPSRC grant EP/P031587/1. MG was supported by EPSRC grants EP/T000414/1, EP/R018413/2, \\ EP/P020720/2, EP/R034710/1, EP/R004889/1, and a Royal Academy of Engineering Research Chair.} 

\vspace{5mm}
\hrule height 0.1cm

\bibliographystyle{bibstyle}

\appendix
\onecolumn

\renewcommand\thesection{S\arabic{section}}
\setcounter{section}{0}
{\sffamily \noindent \huge \bfseries Supporting Information}
\vspace{2cm}
\renewcommand\thefigure{S\arabic{figure}}
\setcounter{figure}{0}
\phantomsection\addcontentsline{toc}{section}{Supporting Information}

\lstset{frame=l,
  backgroundcolor=\color{gray},
  aboveskip=3mm,
  belowskip=3mm,
  showstringspaces=false,
  columns=flexible,
  basicstyle={\small\ttfamily},
  numbers=none,
  numberstyle=\tiny\color{gray},
  keywordstyle=\color{blue},
  commentstyle=\color{dkgreen},
  stringstyle=\color{mauve},
  breaklines=true,
  breakatwhitespace=true,
  tabsize=3
}
\lstdefinestyle{yaml}{
     basicstyle=\color{blue}\footnotesize,
     rulecolor=\color{black},
     string=[s]{'}{'},
     stringstyle=\color{blue},
     comment=[l]{:},
     commentstyle=\color{black},
     morecomment=[l]{-}
 }

\section*{Neural networks}
\subsection*{Notation and Terminology}
A \emph{neural network} is a sequence of length $L \geq 1$ of concatenated transformations. Each \emph{layer} of the net consists of $L_i$ \emph{neurons}, connected through a sequence of \emph{weight matrices} $\mathbf{W}_i~\in~\mathbb{R}^{L_{i+1} \times L_i}$. Each layer applies the transformation
\begin{equation*}
	\sigma_i(\mathbf{W}_i \mathbf{x} + \mathbf{b}_i)
\end{equation*}
to the input $\mathbf{x}$ from the previous layer, where $\mathbf{b}_i \in \mathbb{R}^{L_{i+1}}$ is the \emph{bias} of the $i$-th layer. The function $\sigma_i: \mathbb{R}^{L_{i+1}} \rightarrow \mathbb{R}^{L_{i+1}}$ is the \emph{activation function}; popular choices include the \emph{rectified linear unit} (ReLU) activation function $\sigma(x) = \max(x, 0)$, and the \emph{sigmoid activation function} $\sigma(x) = (1 + e^{-x})^{-1}$. A neural net has an \emph{input layer}, an \emph{output layer}, and \emph{hidden layers}, which are the layers in between the in- and output layers. If a network only has one hidden layer, we call it \emph{shallow}, else we call the neural net \emph{deep}.

\begin{figure}[h]
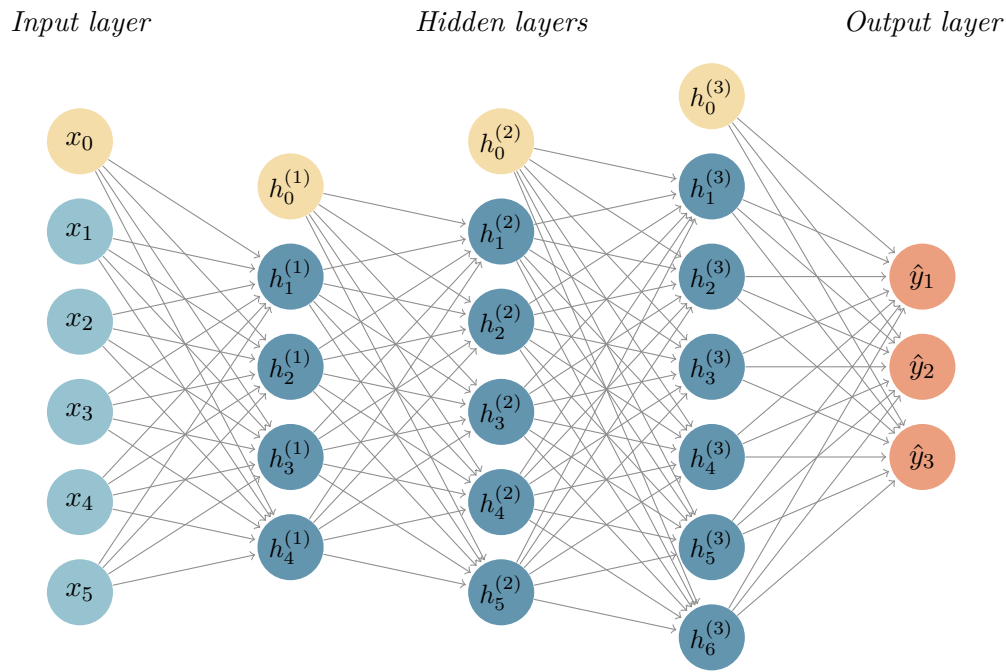

\centering
\definecolor{yellow}{HTML}{F5DDA9}
\definecolor{green}{HTML}{48675A}
\definecolor{red}{HTML}{ec7070}
\definecolor{blue}{HTML}{2F7194}
\definecolor{lightblue}{HTML}{97c3d0}
\definecolor{orange}{HTML}{EC9F7E}
\definecolor{brown}{HTML}{C6BFA2}
 \begin{neuralnetwork}[height=7, nodesize=25, 
 		nodespacing=12mm, layerspacing=28mm]
  		
  		\tikzstyle{bias neuron}=[neuron, fill=yellow!100];
  		\tikzstyle{hidden neuron}=[neuron, fill=blue!75];
  		\tikzstyle{output neuron}=[neuron, fill=orange!100];
  		\tikzstyle{input neuron}=[neuron, fill=lightblue!100];
        
        \newcommand{\x}[2]{$x_#2$}
        \newcommand{\y}[2]{$\hat{y}_#2$}
        \newcommand{\hfirst}[2]{\small $h^{(1)}_#2$}
        \newcommand{\hsecond}[2]{\small $h^{(2)}_#2$}
        \newcommand{\hthird}[2]{\small $h^{(3)}_#2$}
        
        \inputlayer[count=5, bias=true, title={\it Input layer}, text=\x]
        \hiddenlayer[count=4, bias=true, text=\hfirst] \linklayers
        \hiddenlayer[count=5, bias=true, title={\it Hidden layers}, text=\hsecond] \linklayers
        \hiddenlayer[count=6, bias=true, text=\hthird] \linklayers
        \outputlayer[count=3, title={\it Output layer}, text=\y] \linklayers
\end{neuralnetwork}
\vspace{5mm}
\caption{Example of a deep neural network with 3 hidden layers. The inputs (light blue nodes) are passed through the layers, with links between layers representing the weight matrices $\mathbf{W}$. Each layer also applies a \emph{bias} (yellow nodes), with the network finally producing an output (orange).}
\label{diag:neural_net}
\end{figure}

\subsection*{Choice of architecture}
\begin{figure*}[ht!]
\begin{minipage}{0.5\textwidth}
    \cs\flushleft{\textbf{(a)} $L^1$ prediction error}
\end{minipage}
\begin{minipage}{0.5\textwidth}
    \cs\flushleft{\textbf{(b)} Training loss}
\end{minipage}
\includegraphics[width=\textwidth]{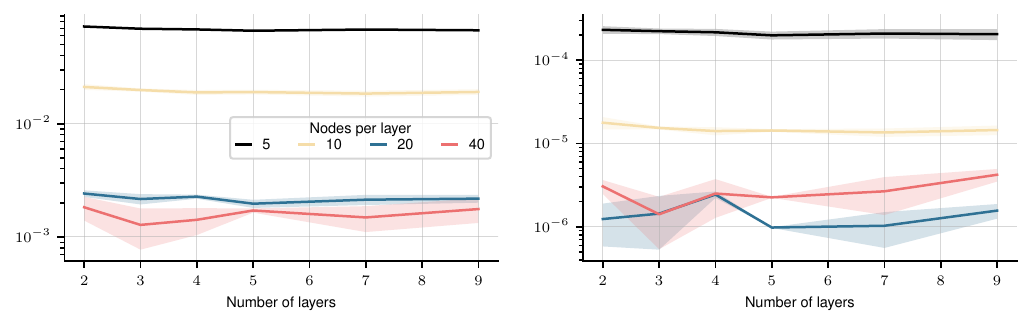}
\begin{minipage}{0.5\textwidth}
    \cs\flushleft{\textbf{(c)} $L^1$ prediction error}
\end{minipage}
\begin{minipage}{0.5\textwidth}
    \cs\flushleft{\textbf{(d)} Avg. compute time per epoch [s]}
\end{minipage}
\includegraphics[width=\textwidth]{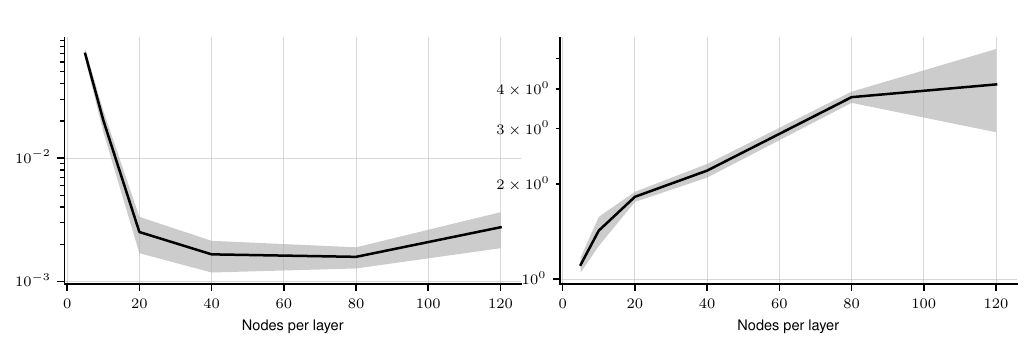}
\caption{Hyperparameter sweep results across neural network depth and width on synthetic Kuramoto data with $N=100$. \textbf{(a)}: the total $L^1$ prediction error eq. [\ref{eq:L1_error}] on the network adjacency matrix. \textbf{(b)}: the training loss $J$, both as a function of the number of layers and the number of nodes per layer in a simple feed-forward architecture. \textbf{(c)}: the $L^1$ prediction error as a function of the number of nodes per layer on a neural network with 5 layers. Increasing the layer size only marginally improves the prediction quality beyond a size of 20, and even decreases it for layer sizes greater than 80. \textbf{(d)}: average compute time per epoch as a function of the layer width.}
\label{fig:hyperparameters_1}
\end{figure*}

Here we provide additional studies to justify our choice of neural architecture. To select an appropriate architecture, we ran a hyperparameter sweep on synthetic Kuramoto data with $N=100$ nodes, 70 training datasets and $L=10$ time steps per dataset. Using a simple feed-forward architecture, we performed a sweep over the number of layers, nodes in each layer and activation functions used. Figures \ref{fig:hyperparameters_1}a—d show that using 5 layers with 20 nodes per layer reduces the training loss $J$ optimally, and that further increasing the layer size only marginally reduces the $L^1$ prediction error. In addition, using very large, highly overparametrised models has the disadvantage of increased computational cost. Similar results hold for the Harris-Wilson dynamics eq. [\ref{eq:HarrisWilson}].

\begin{figure*}[ht!]
\begin{minipage}{0.5\textwidth}
    \cs\flushleft{\textbf{(a)} Deep layer activation functions}
\end{minipage}
\begin{minipage}{0.5\textwidth}
    \cs\flushleft{\textbf{(b)} Final layer activation function}
\end{minipage}
\begin{minipage}{0.5\textwidth}
    \includegraphics[width=\textwidth]{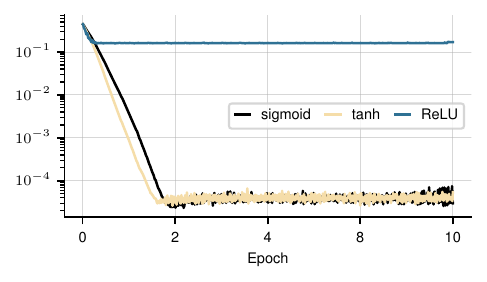}
\end{minipage}
\begin{minipage}{0.5\textwidth}
    \includegraphics[width=\textwidth]{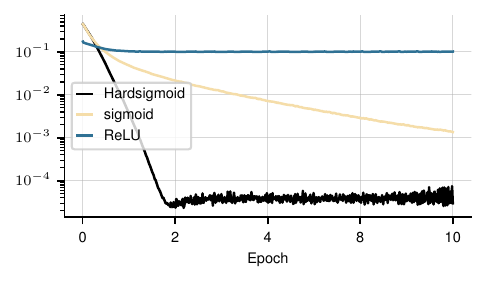}
\end{minipage}
\caption{Hyperparameter sweep results across neural network activation functions on the same data as in fig. \ref{fig:hyperparameters_1}. \textbf{(a)}: the total $L^1$ prediction error on the network adjacency matrix as a function of the activation functions used on the deep layers. On the final layer, the hard sigmoid is used. \textbf{(b)}: the $L^1$ prediction error as a function of the final layer activation function. On the inner layers, the hyperbolic tangent is used.}
\label{fig:hyperparameters_2}
\end{figure*}

We also considered the use of different activation functions on the deep and final layers of the neural network (cf. fig. \ref{fig:hyperparameters_2}). On the inner layers, both a sigmoid and a hyperbolic tangent produce good results, while using anything other than the hard sigmoid on the final layer leads to poor results.

\begin{figure*}[ht!]
\begin{minipage}{0.5\textwidth}
    \cs\flushleft{\textbf{(a)} $L^1$ prediction error}
\end{minipage}
\begin{minipage}{0.5\textwidth}
    \cs\flushleft{\textbf{(b)} Training loss}
\end{minipage}
\begin{minipage}{\textwidth}
    \includegraphics[width=\textwidth]{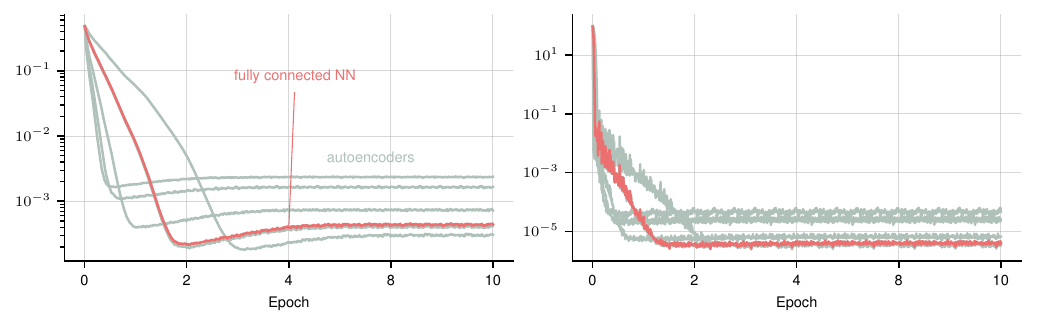}
\end{minipage}
\caption{$L^1$ prediction error \textbf{(a)} and total training loss \textbf{(b)} for various autoencoder architectures with different depths and widths (grey lines) and the simple feed-forward architecture (red) used in this work. The autoencoder architectures range from 25 neurons across 3 layers to about 430 neurons across 11 layers, and synthetic Kuramoto on a network with $N=200$ nodes was used.}
\label{fig:hyperparameters_3}
\end{figure*}

Lastly, using an autoencoder architecture instead of a feed-forward network produces no tangible benefits, so in the interest of simplicity we choose a simple feed-forward architecture \ref{fig:hyperparameters_3}.

\begin{figure*}[ht!]
   \begin{minipage}{0.5\textwidth}
    \cs\flushleft{\textbf{(a)} Distances $l_{ij}$}
\end{minipage}
\begin{minipage}{0.5\textwidth}
    \cs\flushleft{\textbf{(b)} Voltages $U_{ij}$}
\end{minipage}
\begin{minipage}{0.5\textwidth} 
    \includegraphics[width=\textwidth]{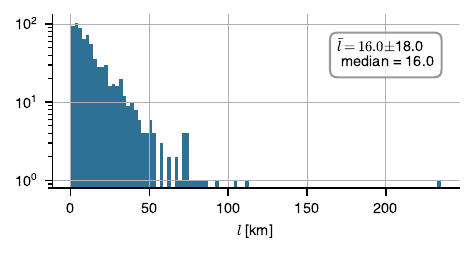}
\end{minipage}
\begin{minipage}{0.5\textwidth} 
    \includegraphics[width=\textwidth]{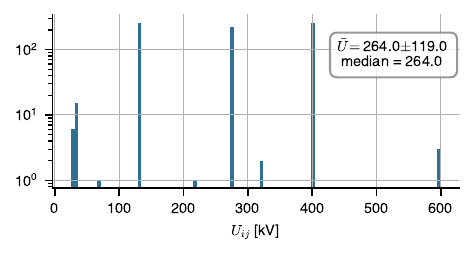}
\end{minipage}
\begin{minipage}{0.5\textwidth}
    \cs\flushleft{\textbf{(c)} Admittances $Y_{ij}$}
\end{minipage} 
\begin{minipage}{0.5\textwidth}
    \cs\flushleft{\textbf{(d)} Scaled edge weights $a_{ij} \sim \vert Y_{ij} \vert U_{ij}^2$}
\end{minipage} 
\begin{minipage}{0.5\textwidth} 
    \includegraphics[width=\textwidth]{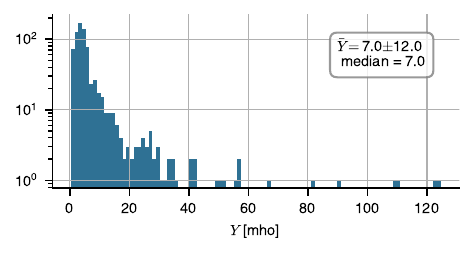}
\end{minipage}
\begin{minipage}{0.5\textwidth} 
    \includegraphics[width=\textwidth]{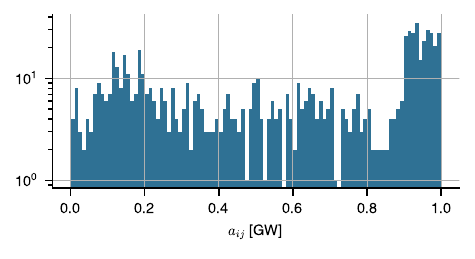}
\end{minipage}
\caption{Line data statistics for the British power grid. \textbf{(a)} Histogram of the node distances in the network, with mean, standard deviation, and median given. \textbf{(b)} Histogram of the line voltages. \textbf{(c)} Histogram of the line admittances. As is visible, a small number of short lines artifically skew the distribution. \textbf{(d)} The resulting normalised edge weights. Short edges with an artificially high admittance are assigned a random value in [0.9, 1].}
\label{fig:edge_weight_statistics}
\end{figure*}

\section*{Inferring line failures in the British power grid}
\subsection*{Initialisation of the neural network weights}
We initialise the neural network's weights with a prior $\pi^0(\boldsymbol{\theta})$ in such a way that the prior $\pi^0(\pred{A})$ is a delta distribution on the complete graph, $\pi(\hat{a}_{ij}) \sim \delta(1) \ \forall i, j$. This can easily be achieved by training the neural network using the simple loss function
\begin{equation*}
    J = \Vert \hat{\mathbf{A}} - \mathbf{1} \Vert_2,
\end{equation*}
with $\mathbf{1} \in \mathbb{R}^{N \times N}$ a matrix of ones in all entries except on the diagonal, where it is zero. This only requires a few training steps, and helps maximise the sampling domain on each edge, making the calculations of p-values in our example more straightforward. However, we should stress that this initialisation is not necessary to obtain good calibration results.

\subsection*{Constructing the admittance matrix}
Here we provide some additional information on the calculation of the network edge weights from the data. The weights quantify the admittance (inverse impedance) of the line, that is, how easily the line can transmit electrical current. The characteristic impedance of a transmission line is given by 
\begin{equation}
    Z_0 = \sqrt{\dfrac{R + i \Omega L}{i \Omega C}},
\end{equation}
with $i$ the complex unit, $\Omega = 50 \ \mathrm{Hz}$ the grid frequency \cite{National_Grid_2023b}, $L$ the cable inductance, and $C$ its capacitance. We use the following values for a copper conductor with a cross-section of 1000 mm$^2$, provided by a standard manufacturer\footnote{\href{https://www.caledonian-cables.co.uk/products/hv/400kv.shtml}{caledonian-cables.co.uk/products/hv/400kv.shtml}} of power grid cables: $R=0.0276$ $\Omega$/km (AC resistance), $L=0.41$ mH/km, $C=0.150$ $\mu$F/km (assuming a single core). The total impedance along the line is then given by
\begin{equation}
    Z(l) = Z_0 \mathrm{sinh}(\gamma l),
\end{equation}
where $l$ is the length of the line, and the propagation constant $\gamma$ is given by
\begin{equation}
    \gamma = \sqrt{(R+i\Omega L)(i \Omega C)}.
\end{equation}
In these calculations we neglect the conductance of the dielectric along the line, which is neglibile at the distances present in the network. The admittance of the line is then given by 
\begin{equation}
    Y_{ij} = Z_{ij}^{-1},
\end{equation}
and the total edge weight $a_{ij}$ by 
\begin{equation}
    a_{ij} = \vert Y_{ij} \vert U_{ij}^2.
\end{equation}
Long lines typically carry two to four times as many cables as shorter lines. We account for this by multiplying the admittance of lines over 10 km by a factor of 2, and those over 80 with a factor of 3 (admittances are additive). Since the neural network outputs values in $[0, 1]$, we scale the edge weights to this range. However, a small number of short lines artificially skew the distribution (see figure \ref{fig:edge_weight_statistics}); these are mainly small segments designed to more accurately capture the geometry of a longer line, though sometimes they represent short lines to transformers, power stations, etc. To reduce their impact on the weight distribution, instead of dividing the weights by the maximum value, we divide by the mean and truncate all weights to $[0, 1]$:
\begin{equation}
    a_{ij} \to \min(1, a_{ij}/\langle a_{ij} \rangle).
\end{equation}
The scaling factor is absorbed into the coupling coefficient $\kappa$. Those edges with weight exactly equal to 1 are reassigned a value chosen uniformly at random in $[0.9, 1.0]$. As is visible in fig. \ref{fig:edge_weight_statistics}d, the resulting distribution of the weights $a_{ij}$ is more uniform than that of the distances $l_{ij}$ and admittances $Y_{ij}$, since longer lines' lower admittance is often compensated by a higher voltage.

These calculations do not account for, among other things, the fact that we are connecting the nodes with straight lines rather than the real line trajectory, are not discerning between overground and underground lines, are assuming a single core per cable, and are not considering differences in transmission line heights, geometries, and materials. These inaccuracies are absorbed into the coefficients $\alpha$ and $\beta$, which we tune manually in order to allow for stable phase-locking. 

\section*{Comparative performance analysis}
\setlength\intextsep{0pt}
\begin{wrapfigure}[16]{r}{0.33\textwidth}
\includegraphics[width=\linewidth]{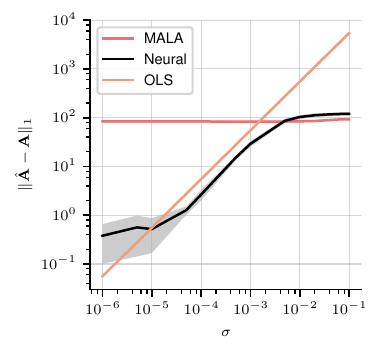}
\caption{$L^1$ prediction error of the neural scheme, the preconditioned Metropolis-adjusted Langevin sampler, and OLS regression as a function of the noise variance $\sigma$ on the training data, for second-order Kuramoto dynamics ($\alpha=1$).}
\label{fig:second_order_accuracy}
\end{wrapfigure}
In figure \ref{fig:second_order_accuracy} we show the equivalent of fig. \ref{fig:computational_performance}a for the case of second-order Kuramoto dynamics (eq. [\ref{eq:second_order_Kuramoto}] with $\alpha=1$). We again see the neural scheme outperforming OLS regression at very low levels of the noise, though the performance improvement is less stark than in the first-order case. MALA marginally outperforms the neural scheme at very high noise levels. Enough data is used to ensure full invertibility of the Gram matrix ($\mathfrak{c}=N-1$).

\section*{Details on the code}
The code is uploaded to the \href{https://github.com/ThGaskin/NeuralABM}{Github repository} as given in the main text. The two models relevant to this work are \texttt{Kuramoto} and \texttt{HarrisWilsonNW}.
\subsubsection*{Installation}
Detailed installation instructions are given in the repository. First, clone the repository, install the \href{https://utopya.readthedocs.io/en/latest/}{\texttt{utopya}} package and all the required additional components into a virtual environment, for example via \texttt{PyPi}. In particular, install \href{https://pytorch.org}{\texttt{pytorch}}. Enter the virtual environment. Then, from within the project folder, register the project:
\begin{lstlisting}
utopya projects register .
\end{lstlisting}
You should get a positive response from the utopya CLI and your project should appear in the project list when calling:
\begin{lstlisting}
utopya projects ls
\end{lstlisting}
Note that any changes to the project info file need to be communicated to utopya by calling the registration command anew. You will then have to additionally pass the \texttt{--exists-action overwrite} flag, because a project of that name already exists. See 
\begin{lstlisting}
utopya projects register --help
\end{lstlisting}
for more information.
Finally, register the project and its models via
\begin{lstlisting}
utopya projects register . --with-models
\end{lstlisting}
\subsubsection*{Running the code}
To run a model, execute the following command:
\begin{lstlisting}
utopya run <model_name>
\end{lstlisting}
By default, this runs the model with the settings in the \texttt{<model\_name>\_cfg.yml} file. All data and the plots are written to an output directory, typically located in \texttt{\textasciitilde/utopya\_output}. To run the model with different settings, create a \texttt{run\_cfg.yml} file and pass it to the model like this:
\begin{lstlisting}
utopya run <model_name> path/to/run_cfg.yml
\end{lstlisting}
This is recommended rather than changing the default settings, because the defaults are parameters that are known to work and you may wish to fall back on in the future. 

Plots are generated using the plots specified in the \texttt{<model\_name>\_plots.yml} file. These too can be updated by creating a custom plot configuration, and running the model like this:
\begin{lstlisting}
utopya run <model_name> path/to/run_cfg.yml --plots-cfg path/to/plot_cfg.yml
\end{lstlisting}
See the \href{https://docs.utopia-project.org/html/getting_started/tutorial.html}{Utopia tutorial} for more detailed instructions.

All the images in this article can be generated using so-called \emph{configuration sets}, which are complete bundles of both run configurations and evaluation configurations. For example, to generate the predictions on the random network with $N=1000$ nodes (fig. \ref{fig:computational_performance}d--e) for the Kuramoto model, you can call
\begin{lstlisting}
utopya run Kuramoto --cfg-set N_1000_example
\end{lstlisting}
This will run and evaluate the \texttt{Kuramoto} model with all the settings from the \texttt{Kuramoto/cfgs/N\_100\_example/run.yml} and \texttt{eval.yml} configurations.
\subsubsection*{Parameter sweeps}
Parameter sweeps are automatically parallelised by \texttt{utopya}, meaning simulation runs are always run concurrently whenever possible. The data is automatically stored and loaded into a data tree. To run a sweep, simply add a \texttt{!sweep} tag to the parameters you wish to sweep over, and specify the values, along with a default value to be used if no sweep is performed:
\begin{lstlisting}
param: !sweep 
  default: 0
  values: [0, 1, 2, 3]
\end{lstlisting}
Then in the run configuration, add the following entry:
\begin{lstlisting}
perform_sweep: true
\end{lstlisting}
Alternatively, call the model with the flag \texttt{-{}-run-mode sweep}. The configuration sets used in this work automatically run sweeps whenever needed, so no adjustment is needed to recreate the plots used in this work.
\subsubsection*{Initialising the neural net}
The neural net is controlled from the \texttt{NeuralNet} entry of the configuration:
\begin{lstlisting}
NeuralNet:
  num_layers: 4
  nodes_per_layer: 
    default: 20
    layer_specific:
      1: 10
      2: 15
  biases:                   # optional; if this entry is omitted no biases are used
    default: ~              # default is None (indicted by a tilde in YAML)
    layer_specific:
      0: default            # use pytorch default (xavier uniform)
      -1: [-1, 1]           # uniform initialisation on a custom interval
  activation_funcs:         
    default: sigmoid   
    layer_specific:         # optional
      1: tanh
      3:
        name: HardTanh     # you can also pass a function that takes additional args and/or kwargs
        args:
          - -2  # min_value
          - +2  # max_value
  learning_rate: 0.001      # optional; default is 0.001
  optimizer: SGD            # optional; default is Adam           
\end{lstlisting}
\texttt{num\_layers} specifies the depth of the net; \texttt{nodes\_per\_layer} controls the architecture: provide a \texttt{default} size, and optionally any deviations from the default under \texttt{layer\_specific}. The keys of the \texttt{layer\_specific} entry should be indices of the layer in question. The optional \texttt{biases} entry determines whether or not biases are to be included in the architecture, and if so how to initialise them. A default and layer-specific values can again be passed. Setting an entry to \texttt{default} initialises the values using the pytorch default initialiser, a Xavier uniform initialisation. Passing a custom interval instead initialises the biases uniformly at random on that interval, and passing a tilde \texttt{\textasciitilde} (\texttt{None} in YAML) turns the bias off. \texttt{activation\_funcs} is a dictionary specifying the activation functions on each layer, following the same logic as above. Any \href{https://pytorch.org/docs/stable/nn.html}{pytorch activation function} is permissible. If a function requires additional arguments, these can be passed as in the example above. Lastly, the \texttt{optimizer} keyword takes any argument \href{https://pytorch.org/docs/stable/optim.html#base-class}{allowed in pytorch}. The default optimizer is the Adam optimizer \cite{Kingma_2014} with a learning rate of $0.02$.

The neural net can be initialised from different initial values in the parameter space by changing the \emph{random seed} in the configuration:
\begin{lstlisting}
seed: 42
\end{lstlisting}
Sweeping over different initialisations is achieved by sweeping over the seed, as described in the previous section.
\subsubsection*{Training the neural net}
The \texttt{Training} entry of the configuration controls the training process:
\begin{lstlisting}
Training:
  batch_size: 2  
  device: cpu
  true_parameters:
    sigma: 0.0
  loss_function:
    name: MSEloss
    # can pass additional args and kwargs here ...
\end{lstlisting}   
You must specify the noise level to use for the ABM during training; the default value is $0$. Under the \texttt{loss\_function} key you can specify the loss function to use, and pass any arguments or keyword arguments it may require using an \texttt{args} or \texttt{kwargs} key. You can use any available \href{https://pytorch.org/docs/stable/nn.html#loss-functions}{pytorch loss function}. 

The \texttt{device} key sets the training device. The default is the CPU, but you can also train on the GPU by setting the device to \texttt{cuda}. Note that on Apple Silicon, the device name is \texttt{mps}. Make sure you have installed the correct package for your device, and note that, as of writing, some pytorch functions required for our code (e.g., the trace and hard sigmoid functions) had not yet been implemented for MPS, hence GPU training on Apple Silicon devices was not possible.
\section*{Kuramoto model}
\subsection*{Neural Network Architecture}
The following is the default configuration for the neural network and training settings used for the Kuramoto model:

\begin{lstlisting}
Kuramoto:
  NeuralNet:
    num_layers: 5
    nodes_per_layer: 
      default: 20
    biases: 
      default: ~    # No biases
    activation_funcs:
      default: tanh                 
      layer_specific:
        -1: HardSigmoid     # hard sigmoid on the last layer
    learning_rate: 0.002 
    optimizer: Adam   
  Training:
    batch_size: 2        
    loss_function:
      name: MSELoss
      kwargs:
        reduction: sum
    true_parameters:
      sigma: 0   
\end{lstlisting}
 
\subsection*{Training}
We rewrite [\ref{eq:second_order_Kuramoto}] as a vector-matrix equation,
\begin{equation*}
	\alpha \dfrac{\mathrm{d}^2 \boldsymbol{\varphi}}{\mathrm{d} t^2} + \beta\dfrac{\mathrm{d} \boldsymbol{\varphi}}{\mathrm{d} t} = \mathbf{P} + \kappa \mathrm{diag}(\mathbf{A} \boldsymbol{\Gamma}(\boldsymbol{\varphi})),
\end{equation*}
which is more amenable to machine learning purposes, since it can make use of fast matrix-multiplication operations. $\boldsymbol{\Gamma}$ is the interaction kernel matrix, $\Gamma_{ij} = \sin(\varphi_j - \varphi_i)$. We the train the model using the following numerical operation: let $\boldsymbol{\varphi}(t) = (\varphi_1(t), ... \varphi_N(t))$ be the current phases of the $N$ nodes, $\boldsymbol{\dot{\varphi}}(t) = (\varphi_1(t)-\varphi_1(t-1), ..., \varphi_N(t)-\varphi_N(t-1)$ the vector of phase derivatives, $\boldsymbol{\omega} = (\omega_1, ..., \omega_N)$ the vector of eigenfrequencies, and $\boldsymbol{\Gamma}(t) = (\sin(\varphi_j(t) - \varphi_i(t))_{ij}$; then in each iteration of the first-order Kuramoto model ($\beta \neq 0$), we do
\begin{equation}
    \boldsymbol{\varphi}(t+1) = \boldsymbol{\varphi}(t) +  \dfrac{1}{\beta}\left( \boldsymbol{\omega}(t) + \mathrm{diag}\left(\mathbf{\hat{A}} \boldsymbol{\Gamma}(t) \right) \right) \mathrm{d}t,
\end{equation}
where $\mathrm{diag}(\cdot)$ takes the diagonal elements of the matrix. For the second-order model ($\alpha \neq 0$), we do
\begin{equation}
    \boldsymbol{\varphi}(t+1) = \boldsymbol{\varphi}(t) + \dfrac{1}{\alpha}\left[ \left(\boldsymbol{\omega}(t) + \mathrm{diag}\left(\mathbf{\hat{A}} \boldsymbol{\Gamma}(t) \right) - \beta \boldsymbol{\dot{\varphi}}(t) \right) \mathrm{d}t + \boldsymbol{\dot{\varphi}}(t) \right]\mathrm{d}t.
\end{equation}
In the first-order case, the initial conditions required are $\boldsymbol{\varphi}(0)$, in the second order case, we need $\boldsymbol{\varphi}(0)$ and $\boldsymbol{\varphi}(1)$ (i.e. the initial phases and initial velocities).

\subsection*{Running the code}
Configuration sets to reproduce all numerical experiments except the British power grid inference are provided; for instance, to produce the predictions for a random network with $N=100$ nodes, simply run
\begin{lstlisting}
utopya run Kuramoto --cfg-set N_1000_example
\end{lstlisting}

\section*{Harris-Wilson model}

\subsection*{Neural Network Architecture}
The following is the default configuration for the neural network and training settings used for the Harris-Wilson model:

\begin{lstlisting}
HarrisWilsonNW:
  NeuralNet:
    num_layers: 2
    nodes_per_layer: 
      default: 20
    biases: 
      default: ~    # No biases
    activation_funcs:
      default: tanh                 
      layer_specific:
        -1: sigmoid     # sigmoid on the last layer
    learning_rate: 0.002   
    optimizer: Adam           
  Training:
    batch_size: 2
    loss_function:
      name: MSELoss
      kwargs:
        reduction: sum
    true_parameters:
      sigma: 0   
\end{lstlisting}

\subsection*{Training}
We use the following matrix form of the Harris-Wilson equations to train the neural net. Let $\mathbf{D} \in \mathbb{R}^M$, $\mathbf{O} \in \mathbb{R}^N$, $\mathbf{W} \in \mathbb{R}^M$ be the demand vector, origin zone size vector, and destination zone size vector respectively. The the dynamics are given by
\begin{equation}
	\mathbf{D} = \mathbf{W}^\alpha \odot \left[\left(\mathbf{C}^\beta\right)^\top (\mathbf{O} \odot \mathbf{Z})  \right] \in \mathbb{R}^M,
\end{equation}
with $\odot$ indicating the Hadamard product, and elementwise exponentiation. $\mathbf{Z} \in \mathbb{R}^N$ is the vector of normalisation constants
\begin{equation}
	\mathbf{Z}^{-1} =  \mathbf{C}^\beta \mathbf{W}^\alpha.
\end{equation} 
The dynamics then read
\begin{equation}
	\dot{\mathbf{W}} = \mathbf{W} \odot \epsilon(\mathbf{D} - \kappa \mathbf{W})
	\label{eq:matrix_formulation}
\end{equation}
with given initial conditions $\mathbf{W}(t=0) = \mathbf{W}_0$.
This formulation is more conducive to machine learning purposes, since it contains easily differentiable matrix operations and does not use for-loop iteration.
\subsection*{Running the code}
This is analogous to the Kuramoto case. To reproduce the plot from the main article, run the following command:
\begin{lstlisting}
utopya run HarrisWilsonNW --cfg-set London_dataset
\end{lstlisting}

\end{document}